\documentclass{CVIS}
\usepackage{amsmath,bm}
\usepackage{graphicx}
\usepackage{dcolumn}
\usepackage{caption}
\usepackage[numbers,sort&compress,sectionbib]{natbib}
\usepackage[pagebackref=false,breaklinks=true,letterpaper=true,colorlinks=true,linkcolor = blue, urlcolor = black,citecolor=black,bookmarks=true]{hyperref}
\usepackage{url}
\usepackage{mleftright}
\usepackage{multirow}
\usepackage{makecell}
\usepackage{titlesec}
\usepackage[redeflists]{IEEEtrantools}
\usepackage{blindtext, subfig}
\usepackage{dblfloatfix}
\usepackage{algpseudocode}
\usepackage{algorithm}
\usepackage{caption}

\algdef{SE}[SUBALG]{Indent}{EndIndent}{}{\algorithmicend\ }%
\algtext*{Indent}
\algtext*{EndIndent}
\title{Plankton-FL: Exploration of Federated Learning for Privacy-Preserving Training of Deep Neural Networks for Phytoplankton Classification}

\begin{document}
\bstctlcite{IEEEexample:BSTcontrol}
\sloppy
\author{
\begin{tabularx}{\textwidth}{X X}
Daniel Zhang &  University of Waterloo\\
Vikram Voleti & Mila, University of Montreal\\
Alexander Wong & University of Waterloo, Blue Lion Labs\\
Jason Deglint & University of Waterloo, Blue Lion Labs\\
\multicolumn{2}{l}{Email: \{daniel.zhang1, jdeglint\}@uwaterloo.ca}
\end{tabularx}
}

\maketitle
\begin{abstract}
\vspace{-1.65mm}
Creating high-performance generalizable deep neural networks for phytoplankton monitoring requires utilizing large-scale data coming from diverse global water sources. A major challenge to training such networks lies in data privacy, where data collected at different facilities are often restricted from being transferred to a centralized location. A promising approach to overcome this challenge is federated learning, where training is done at site level on local data, and only the model parameters are exchanged over the network to generate a global model. In this study, we explore the feasibility of leveraging federated learning for privacy-preserving training of deep neural networks for phytoplankton classification. More specifically, we simulate two different federated learning frameworks, federated learning (FL) and mutually exclusive FL (ME-FL), and compare their performance to a traditional centralized learning (CL) framework. Experimental results from this study demonstrate the feasibility and potential of federated learning for phytoplankton monitoring.

\end{abstract}
\vspace{-7.8mm}
\section{Introduction}
\vspace{-3mm}
The uncontrollable growth of particular phytoplankton and algae species can cause the formation of harmful algae blooms (HABs). If not properly monitored and controlled, HABs can have severe, negative impacts on various industries, natural ecosystems, and the environment~\cite{inbook}.
HABs are a growing concern as research has shown that climate change has led to an increase in the frequency and severity of HABs~\cite{WELLS2020101632}. A very important step in the monitoring and controlling of HAB formation is the identification of phytoplankton and algae species. Unfortunately, this process is largely manual and thus is highly time-consuming and prone to human error. As such, effective methods for automating the species identification process are highly desired.

Recent advances in machine learning, in particular deep learning, have shown considerable promise for monitoring and assessment of phytoplankton and algae~\cite{deglint2019investigating, Bamra2022TowardsGL}. However, a significant bottleneck to training such models is the need for large-scale data coming from different water sources across different countries in order to create high-performance, generalizable models. Since the data collected at the different facilities are often restricted from being transferred to a centralized location for training due to data privacy concerns, this makes it infeasible to leverage traditional, centralized learning frameworks for building such models.  

A particularly promising direction for tackling this data privacy challenge lies in federated learning (FL), which involves training local models at individual local nodes on the premises (prem) of each local data source and communicating only the parameters and updates of these local models to a server for generating a global model to reap the benefits from the different local data without having seen any of the individual data sources~\cite{McMahan2017CommunicationEfficientLO}. FL has demonstrated considerable success in the domains of mobile computing~\cite{McMahan2017CommunicationEfficientLO, Zhao2018FederatedLW} and healthcare~\cite{Rieke2020}, and thus can hold considerable potential for the application of phytoplankton monitoring and assessment.

\begin{figure}[h]
	\begin{center}
	\includegraphics[width=0.9\columnwidth]{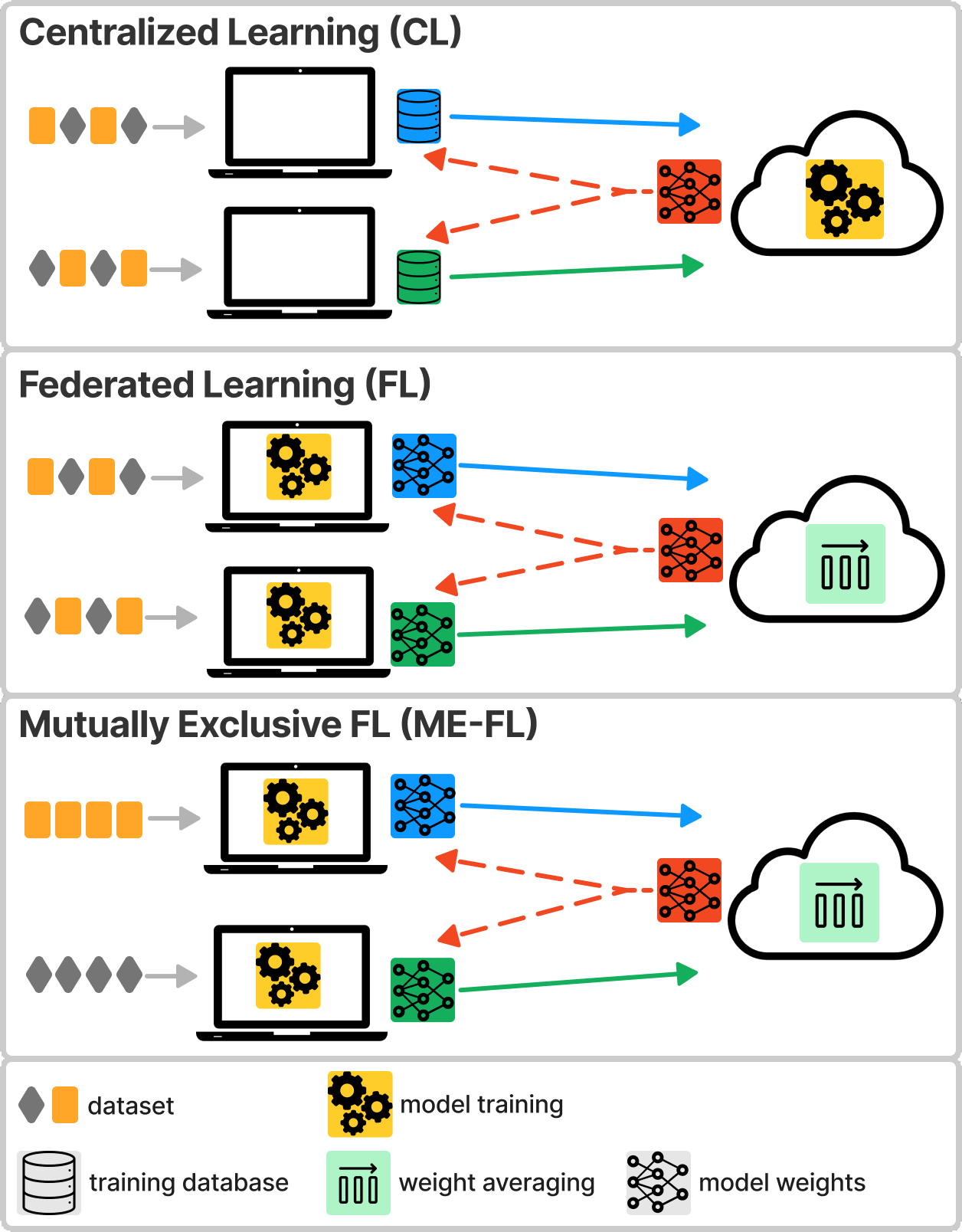}
        \vspace{-6mm}
	\end{center}
	\setlength{\belowcaptionskip}{-12pt}
	\caption{Traditional centralized learning (CL) (top) and the two federated learning frameworks as realizations of Plankton-FL: federated learning (FL) (middle) and mutually exclusive FL (ME-FL) (bottom). Both privacy-preserving federated learning frameworks are evaluated against CL for the training of deep neural networks for phytoplankton classification.}
	\label{fig1}
\end{figure}

In this study, we explore the feasibility of leveraging federated learning to train deep, convolutional neural networks for the purpose of image-driven phytoplankton classification, which we will refer to as Plankton-FL. Our main contributions in this study are as follows: (1) we simulate and study two federated learning frameworks as potential realizations of Plankton-FL: (centralized) federated learning (FL) and mutually exclusive FL (ME-FL), (2) we evaluate the performance of both Plankton-FL frameworks, and (3) we compare them to a traditional, centralized learning framework (CL). Figure \ref{fig1} provides a visual representation of each of the three environments.

\section{Methodology}

\subsection{Background}

Federated learning has been shown to be very effective for training deep neural networks on decentralized data while ensuring data privacy~\cite{McMahan2017CommunicationEfficientLO}. Specifically, when there is sensitive data from various sources, federated learning can be leveraged. A typical federated learning framework consists of 2 components: a global model and $K$ clients. Each client contains its own local model, and they are trained iteratively and independently on their respective data. It is assumed that all of the data available is partitioned into $K$ clients, $P_k$. The local models are then used to update the global model~\cite{McMahan2017CommunicationEfficientLO}. This process is repeated for $N$ rounds in order for the global model to generalize. The objective for federated learning is calculated using equation~\ref{eq1}.

\begin{equation}
f(w) = \sum_{k = 1}^{K}\frac{n_k}{n}F_k(w) \quad \textrm{where} \quad F_k(w) = \frac{1}{n_k}\sum_{i \in P_k}f_i(w)
\label{eq1}
\end{equation}

In equation~\ref{eq1}, $f_i(w)$ denotes the loss function $\ell(x_i, y_i; w)$ for an observation $(x_i, y_i)$ and model parameters $w$. Also, $n_k$ and $n$ denote the $|P_{k}|$ and the total number of observations, respectively.

The centralized federated learning algorithm which governs the communication between the global and local models is known as FederatedAveraging (FedAvg) and was introduced by McMahan \textit{et al.} \cite{McMahan2017CommunicationEfficientLO}. FedAvg is the iterative process of training all the local models, taking the average of all the updated weights from the local models, and then using it to update the global model. As described by McMahan \textit{et al.}, pseudo-code is provided in Algorithm 1.
\begin{algorithm}
\centering
    \caption{McMahan \textit{et al.} \cite{McMahan2017CommunicationEfficientLO}'s implementation of the FedAvg algorithm. $C$ is the set of all clients; $B$ is the local model batch size; $LE$ is the number of local epochs; $\eta$ is the learning rate; $w$ are the weights, and $\ell$ is the loss function}\label{algorithm}
    \begin{algorithmic}[1]
        \State \textbf{Global/Server:}
        \State $C\gets\text{Set of all available clients}$
        \State \text{initialize $w_0$}
        \For{each round, $r = 1$,$2$, ...} 
        \For{each client, $k \in C$,}
            \State $w_{r+1}^k \gets \textbf{Local-Training($k$, $w_r$)}$
            \EndFor
        \State \text{FedAvg:} $w_{r+1}\gets\sum_{k=1}^K \frac{n_k}{n}w^k_{r+1}$
        \State \text{SYNC Global}$\rightarrow$\text{Local: } $w_{r+1}^k \gets w_{r+1} \ \ \forall k$
        \EndFor 
        \\
        \State \textbf{Local-Training($k$, $w$):}
        \Indent
        \State \text{Split data on client $k$ into $B$ batches}
        \For{each local epoch, $l = 1...LE$}
        \For{batch $b \in B$}
        \State $w \gets w - \eta * \nabla_w\ \ell(w ; b)$
        \EndFor
        \EndFor
        \State \text{return $w$}
        \EndIndent
    \end{algorithmic}
\end{algorithm}



In this paper, we used the FedAvg method, as described in Algorithm 1, when training our two instances of Plankton-FL.
To test the feasibility and potential of federated learning, three different experiments were simulated. Specifically, a centralized learning baseline (CL), a (centralized) federated learning framework (FL), and a mutually exclusive, federated learning framework (ME-FL). Figure \ref{fig1} provides a visual representation of each of the three experiments. 

\subsection{Centralized Learning (CL)}
For the CL experiment, we have two data sources that we consolidate into a single server. From there we train a centralized model, which can then be deployed back to the edge devices for assessment and monitoring. The model was trained for a maximum of 75 epochs, with an early stopping criteria: A minimum of 50 epochs and a $\delta$ between test accuracies of 0.000001.  

\subsection{Federated Learning (FL)}

In the FL experiment, all of the training data was combined, randomly shuffled, and distributed to clients. Each client trained their own local model, on-prem, and only communicated their parameters back to the global server. FL was run for 10 iterations, where each iteration number corresponded to the number of clients. Namely, for the first iteration, there was only one client containing all of the training data, identical to CL, and with each increasing iteration another client was added (i.e. second iteration utilized two clients, etc.). Although, in reality, no single client will contain all of the data, for the purpose of comparing to CL, it made sense to start with a single client. 

\subsection{Mutually Exclusive FL (ME-FL)}

Unlike FL, in ME-FL, instead of combining all of the data together, shuffling, and distributing them, the clients only contained data from a single source, making them mutually exclusive. Again, each of the clients trained their own model, on-prem, and only communicated their parameters back to the global server. ME-FL was run for 9 iterations, starting from 2 clients up to 10. Iterations start from 2 clients due to the nature of the experiment; since each client only has data from a single source, it would not make sense to only have a single client. This modified setup ensures that we always have data from both sources.

\vspace{-2.5mm}

\section{Experimental Setup}

\subsection{Dataset}
\vspace{-1.5mm}
The dataset was provided by Blue Lion Labs and was collected from two mutually exclusive sources, Halifax and Waterloo.
It contained 301 distinct microscope specimen photos, each at a resolution of 3208 x 2200 pixels. The phytoplankton contained in these images are from eleven different species: \textit{Entomoneis paludosa}, \textit{Alexandra catenella}, \textit{Pymnesium parvum}, \textit{Navicula sp}, \textit{Heterosigma akashiwo}, \textit{Prorocentrum lima}, \textit{Alexandrium ostenfeldii}, \textit{Porphyridium purpureum}, \textit{Dolichospermum}, \textit{Phaeodactylum tricornutum M1}, and \textit{Phaeodactylum tricornutum M2}.

\vspace{-3mm}
\subsection{Model Architecture}
For the purpose of this exploration, we took the majority class present in each image as the label to do image classification. This ensures that all models receive the same amount of information. Given the task, we built a custom convolutional neural network with four convolutional layers, three max-pooling layers, two dense layers, and an output layer. Across all intermediate layers, the ReLU activation function was used, and at the output, a softmax activation was used to predict the probabilities of each class. For all the convolutional layers, a kernel of size 3x3 and a stride of 1x1 was used and for all of the pooling layers, a pool and stride size of 2x2 was used. Additionally, dropout was used with a rate of 0.25 after the convolutional layers and a rate of 0.5 after the first dense layer.

\vspace{-1.5mm}
\subsection{Model Training and Evaluation}

When training, the images were resized to a resolution of 128 x 128 pixels and further augmented, using a horizontal flip, vertical flip, rotation, and color jitter, to create a larger data set of 2107 images. Across all experiments, the model architectures were held the same, a batch size of 8 was used, and the data was split into 80\% training and 20\% test. We also tune the learning rate across all experiments via a grid search over three different learning rates (LR) of 0.001, 0.0001, and 0.0005. Both of the federated learning experiments were run for 75 rounds and each local model was trained for 1 epoch. Furthermore, given it is a multi-label image classification task, the metric considered is prediction accuracy and the loss function is categorical cross-entropy.

\vspace{-1.25mm}

\begin{table}
\begin{tabular}{|c|c|c|c|} 
\hline
\textbf{\shortstack{\# of \\clients}} & \textbf{\shortstack{Centralized \\Learning (CL)}} & \textbf{\shortstack{Federated \\Learning (FL)}} & \textbf{\shortstack{Mutually Exclusive \\FL (ME-FL)}} \\
\hline
1 &  & 92\%   & - \\ \cline{1-1}\cline{3-4}
2 & & 89\% & 63\%\\\cline{1-1} \cline{3-4}
3 & & 79\%& 60\%\\\cline{1-1}\cline{3-4}
4 & &75\% &65\%\\\cline{1-1}\cline{3-4}
5 &  \multirow{1}{2em}{91\%} & 82\%&31\% \\ \cline{1-1} \cline{3-4}
6 & &77\% &55\%\\\cline{1-1} \cline{3-4}
7 & & 76\%&47\%\\\cline{1-1} \cline{3-4}
8 & & 78\%&18\%\\\cline{1-1} \cline{3-4}
9 & & 76\%&53\%\\\cline{1-1} \cline{3-4}
10 & & 75\%&16\% \\\cline{1-1} \cline{3-4}
\hline
\end{tabular}
  \centering
  \setlength{\belowcaptionskip}{-10pt}
\caption{The experimental results across all setups, for a learning rate of 0.0001. FL yielded a higher test accuracy than CL for a single client, but for all other clients, it is outperformed. ME-FL consistently yields the lowest test accuracies across every number of clients.}
\label{table1}
\end{table}

\vspace{-2.1mm}

\section{Results \& Discussion}
\vspace{-2.1mm}
\begin{figure}[t!]
   \centering
	\begin{center}
	\includegraphics[width=\columnwidth]{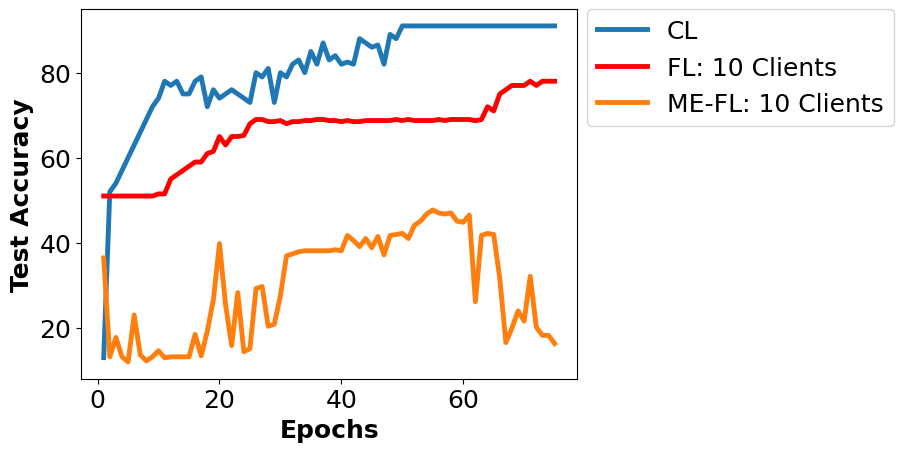}
        \vspace{-7mm}
	\end{center}
	  \setlength{\belowcaptionskip}{-17pt}
	\caption{The test accuracies, across all experiments, per epoch with a learning rate of 0.0001. CL (left) was trained for 51 epochs and achieved a test accuracy of 91\%, both FL (middle) and ME-FL had 10 clients and achieved test accuracies of 78\% and 16\%, respectively after 75 epochs.}
	\label{fig2}
\end{figure}

\subsection{Comparison of Performance Across Experiments}
Table \ref{table1} provides a numerical comparison across all experiments utilizing a learning rate of 0.0001. Note that, CL, FL, and ME-FL were trained for 51, 75, and 75 epochs, respectively. Firstly, when comparing CL and FL, we observe that for a single client FL outperforms CL. However, this is expected because FL with a single client is the exact same setup as CL; we expect the test accuracies to be very close in magnitude, and it is entirely possible that FL can outperform CL in this scenario. For all other number of clients, FL has a progressively worse test accuracy and is continuously outperformed by CL. In addition, across all number of clients, we observe that ME-FL consistently gets outperformed by CL and FL, which further demonstrates the impracticality of this method. 

Figure \ref{fig2} provides a visual comparison of CL, FL, and ME-FL across all epochs. We specifically look at the results for 10 clients of both FL and ME-FL, as in reality there are often large numbers of clients. From the figure, CL and FL both appear to learn, whereas ME-FL does not appear to learn at all. Comparing CL and FL, we observe that CL converges much faster than FL, which tells us that CL learns faster than FL.  
Overall, across all experiments, generally, we observed that CL performed the best, FL performed the second best, and ME-FL had the worst performance and this same trend is observed across each learning rate.

\vspace{-1.5mm}
\subsection{Downward Trend in FL Across Number of Clients}
\begin{figure}[t!]
	\begin{center}
	\includegraphics[scale = 0.34]{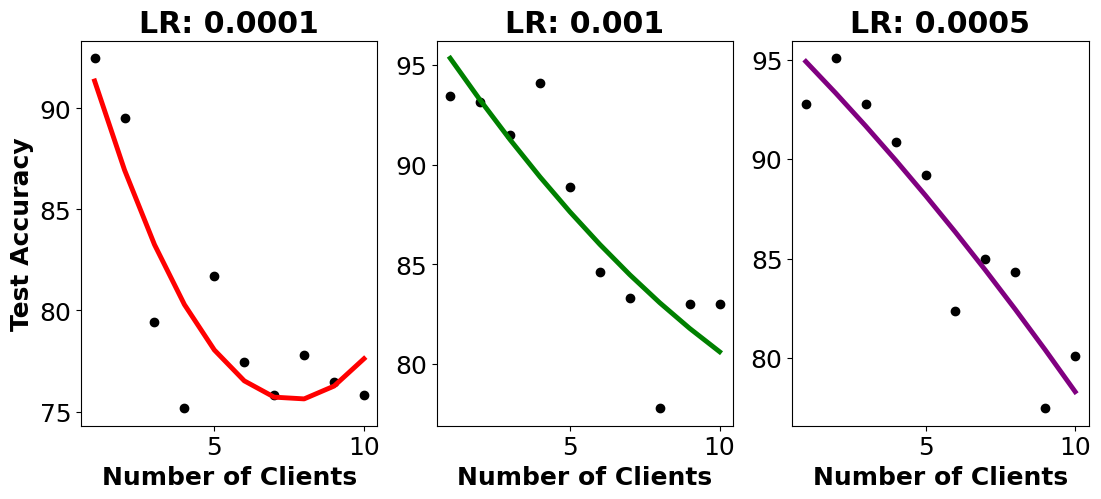}
   \vspace{-6mm}
	\end{center}
	  \setlength{\belowcaptionskip}{-18pt}
	\caption{FL global model final test accuracy per number of clients, under each different learning rate. This illustrates that for all learning rates, we observe a downward trend as we increase the number of clients}
	\label{fig3}
\end{figure}

From table \ref{table1} and figure \ref{fig2}, we observe that FL performs relatively well, which prompted an investigation into its properties. Figure \ref{fig3} displays the global model test accuracies for FL, for all numbers of clients, across the three learning rates. We observe a downward trend in the test accuracies as the number of clients increases. With an increasing number of clients, the global model needs to process and learn more information (i.e. the global model has to aggregate weights from more sources). With more information to process, learning is slowed down, yielding a worse generalization. 

\subsection{Causation of Poor ME-FL Performance}

The largest contributor as to why ME-FL had a subpar performance relative to FL was because ME-FL was trained on individual, mutually exclusive clients. In our FL experiments we utilize a homogeneous model architecture, that is, all clients and the global server have the same model architecture. The nature of FL yields an independent and identically distributed (IID) distribution of labels across clients. However, in ME-FL, the distribution across clients is non-IID. As discussed in other literature, homogeneous federated learning performs poorly on non-IID data distributions~\cite{Zhao2018FederatedLW}. 
Given this limitation, heterogeneous federated learning ~\cite{Li2019FedMDHF, Yu2020HeterogeneousFL} is an alternative approach that should be explored. In this method, clients are allowed to differ in network architecture, allowing for more flexibility. Research has been done to explore applications of heterogeneous federated learning to mutually exclusive data~\cite{Gudur2020FederatedLW} and it has typically been the preferred approach over homogeneous federated learning.

\section{Conclusion \& Future Works}

This work demonstrates the feasibility and potential of Plankton-FL for the privacy-preserving building of high-performance, generalizable models for phytoplankton assessment without the need to exchange data. We simulated two different federated learning frameworks and compared their performance to a traditional, centralized learning framework. Although centralized learning yields the best performance, it does not address privacy concerns. Federated learning preserves privacy but fails to generalize when clients are mutually exclusive. We find that when clients share class labels with one another, federated learning both generalizes well and provides a privacy-preserving alternative to centralized learning. 

Given the outcomes of this paper, the immediate future work includes (1) implementing this framework for object detection to build off the current work of image classification, (2) utilizing a heterogeneous federated learning framework and conducting the same experiments to assess the relative performance to homogeneous federated learning, and (3) explore novel federated learning-related methods. For example, another method that can be utilized is git re-basin, which aims to train individual models on disjoint datasets and merge them together ~\cite{Ainsworth2022GitRM}. Finally, careful consideration must be taken on how federated learning frameworks will be deployed in the field to ensure data privacy between clients. This will help provide a secure and accurate method for identifying different species of phytoplankton and help alleviate the manual workload. 
\vspace{-1.5mm}

\section*{Acknowledgments}
This work was funded by the Waterloo AI Institute and Mitacs. The dataset was provided by Blue Lion Labs, and the computing resources were provided by the Vision and Image Processing (VIP) Lab at the University of Waterloo and Blue Lion Labs.
\vspace{-1.5mm}
\bibliographystyle{IEEEtran}
\bibliography{references.bib}

\end{document}